# Evaluation Framework for Computer Vision-Based Guidance of the Visually Impaired

Krešimir Romić, Irena Galić, Marija Habijan, Hrvoje Leventić
Faculty of Electrical Engineering, Computer Science and Information Technology Osijek,
Kneza Trpimira 2B, HR-31000 Osijek, Croatia
kresimir.romic@ferit.hr

*Abstract*—Visually impaired persons have significant problems in their everyday movement. Therefore, some of our previous work involves computer vision in developing assistance systems for guiding the visually impaired in critical situations. Some of those situations includes crosswalks on road crossings and stairs in indoor and outdoor environment. This paper presents an evaluation framework for computer vision-based guiding of the visually impaired persons in such critical situations. Presented framework includes the interface for labeling and storing referent human decisions for guiding directions and compares them to computer vision-based decisions. Since strict evaluation methodology in this research field is not clearly defined and due to the specifics of the transfer of information to visually impaired persons, evaluation criterion for specific simplified guiding instructions is proposed.

## I. Introduction

Novel assistance systems for the visually impaired persons can include computer vision methods as the basis for providing guiding instructions for the user. In that particular case, person wearing camera and portable processing device can obtain real-time instructions in movement. This is particularly important in critical situations, e.g., when person is crossing the road on designated crosswalk or when approaching the stairs. Systems where the camera input image presents what is visible from the user's perspective are called first person vision (FPV) systems [1]. In order to provide guiding instructions, in FPV systems usual approach is to: (1) detect region of image where person should be directed; (2) calculate the direction angle; (3) transform direction angle into simplified instruction for the visually impaired person. This approach is shown in Fig. 1. Proposed framework has the main aim to simplify the evaluation process of methods that use this kind of approach. This approach is similar to one in [2] where turn-based navigation is performed and generally to all approaches where turning directions are given relative to user orientation [3], [4]. While the methodology for evaluating ROI detection is well known and covered in this research field, the methodology for evaluating the simplified guiding instructions is not strictly defined and approaches vary. Therefore, this paper suggests evaluation pipeline and criterion integrated in the framework. This framework, along with the database of videos and images should help in future evaluation and comparison to other similar work.

### A. Background

Some of our previous work includes novel methods for detection of crosswalks and stairs in assistance systems for the visually impaired persons [5]. Similarly, some authors address the detection of other obstacles in such assistance systems [6], [7]. Furthermore, those methods are broaden with sound guiding technique to help the users in their movement [8]. However, the evaluation of aforementioned methods became demanding due to the specifics of the application. Therefore, evaluation framework is presented in this paper, along with the new evaluation criterion. This criterion is tailored for approaches where visually impaired person is guided with simplified instructions converted from detected direction angle as shown in Fig. 1.

This paper is organized as follows: Section II presents details of proposed evaluation framework; Section III presents the new criterion for evaluation and the conclusion is given in Section IV.

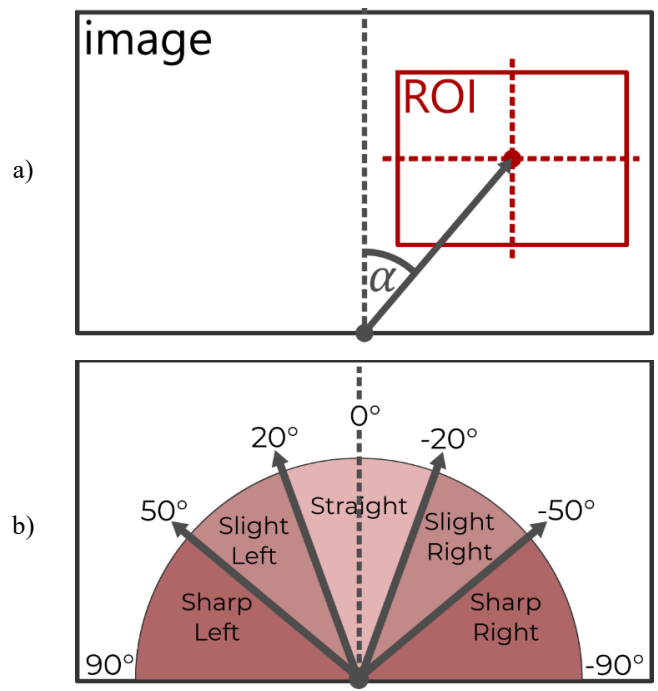

Figure 1. Computer vision-based approach for guiding the visually impaired: a) Find ROI and calculate direction angle α, b) Transform angle into simplified instruction

## II. EVALUATION FRAMEWORK

Developed evaluation framework has the main aim to easily compare human and computer decisions about guiding the visually impaired. That comparison allows proper analysis and evaluation of this kind of methods for guiding. Evaluation framework is described and presented in this section.

### A. Test environment

Evaluation framework has simple test environment where images or videos can be loaded. For every image or video frame, human decisions about guiding the visually impaired can be stored. Those human decisions will be considered as ground truth for evaluation. First type of human decision is to mark the rectangular region of interest on image, e.g., free area on crosswalk or stairs. Second type of decision is to simply mark one of the five possible directions (Fig. 2).

Test environment also contains another part where computer decisions with previously developed methods can be obtained, analyzed, and compared to human decisions for the particular image.

### B. Test data

As mentioned earlier, this research was conducted on two examples of critical situations: crosswalks and stairs. Private database of video sequences for both situations was collected. Frames were extracted from those video sequences with a total of 500 images of crosswalks and 500 images of stairs. Those images include specific scenes with variety of crosswalk/stairs positions on image. Except various positions of the desired objects, database also includes images where crosswalk or stairs are partially occluded.

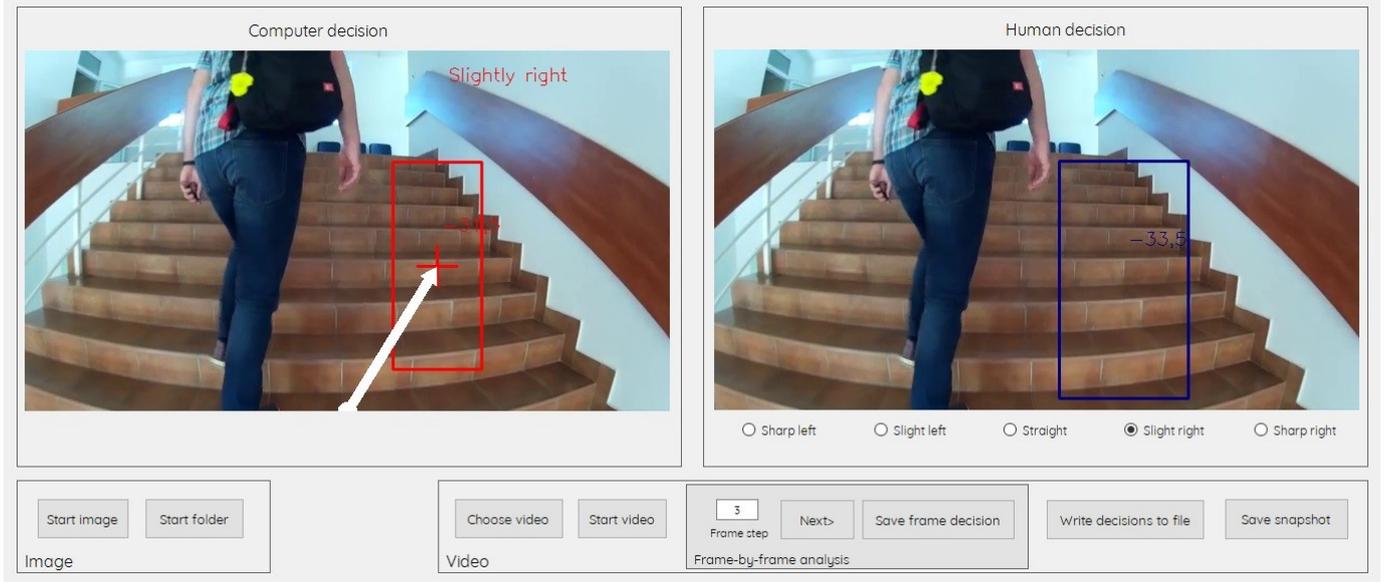

Figure 2. Test environment

### C. Evaluating the accuracy of decisions

Evaluation process can be conducted in two ways. First one is based on the comparison of direction angles. Direction angles are often referred to as azimuth angles [9]. Direction angle obtained by a computer need to be compared with ground truth direction angle and deviation ($\delta$) between two angles is calculated (Fig. 3). In this case, smaller deviations of direction angles will be considered as better results. For example, our database of crosswalk images has an average deviation of 5.99° when using the method presented in [5]. It is similar with stairs database where average deviation of 6.93° was obtained. Furthermore, this angle deviations ($\delta$) can be distributed in ranges in order to analyze the number of significant errors in detection (Fig. 4).

However, final decision and instruction for the visually impaired person will not be presented as an angle. Instructions will be obtained by converting an angle into one of the five directions (Fig. 1 b) and eventually passed to user via sound signals or vibrations [10]. The second way of evaluation is based on comparing ground truth decisions with simplified directions converted from calculated angles. In this case, when having five possible directions, human and computer decisions can be the same or they can deviate in 1-4 levels (e.g. "straight" and "slight left" is one level deviation). Tab. 1 shows the example results for the crosswalk and stairs database.

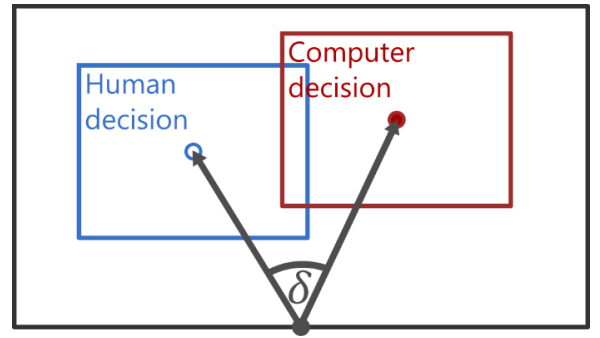

Figure 3. Direction angle deviation ($\delta$) between human and computer decision

As shown in Tab. 1, most of the computer and human decisions are the same (i.e., correct) and none of the decisions deviate more than one level. However, significant number of decisions has one level deviation (i.e., error) and Tab. 1 presents detailed percentages for that case. By analyzing these percentages, it can be concluded that most of the one level errors has direction angle deviation smaller than 15°. Therefore, it is possible that simplified decision was on the boundary between two decisions and one level deviation was recorded even though the calculated angle is near ground truth. In order to have more fair evaluation for this kind of comparison, a new criterion is proposed.

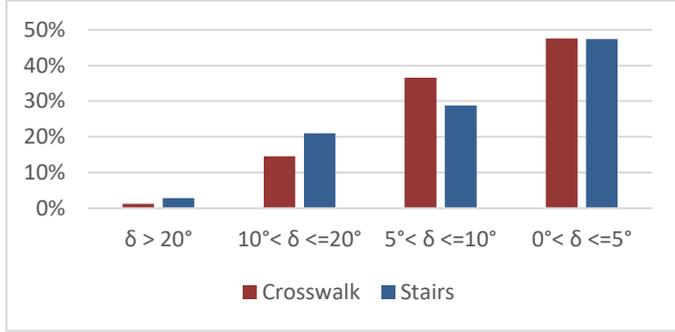

Figure 4. Distribution of direction angle deviations

TABLE I. DISTRIBUTION OF DEVIATION LEVELS FOR GUIDING ON CROSSWALKS AND STAIRS

| Levels of deviation | | | Percentage of decisions | |
|---|---|---|---|---|
| | | | Crosswalk | Stairs |
| Without deviation | | | 89.8 % | 87.0 % |
| 1 level | Overall | | 10.2 % | 13.0 % |
| | deviation < 15° | | 9.4 % | 7.8 % |
| | deviation ≥ 15° | | 0.8 % | 5.2 % |
| 2 levels | | | 0 % | 0 % |
| 3 levels | | | 0 % | 0 % |
| 4 levels | | | 0 % | 0 % |

III. NEW CRITERION FOR EVALUATION

The new criterion for evaluation takes into account the proximity of the direction angle to boundaries between two decisions. According to the new criterion, the decision accuracy will not be exclusively a binary value (0 or 1). In the boundary areas the accuracy will be represented as values between 0 and 1. Thus, the decisions based on an angle that is close to the ground truth will score better accuracy than those based on angle that is far from the ground truth. For example, when a human decision is "sharp left" (reference range of angles from 50° to 90°), and the algorithm calculates the angle of 49.9°, then it can be said that accuracy is high even though the computer decision is "slight left". On the other hand, when the algorithm calculates the angle of 25°, which is also the "slight left" decision, the accuracy of a such decision will be low because the obtained angle is far from the boundary of the ground truth.

To determine the accuracy ($g$) in the boundary areas, the criterion uses a modified Gaussian function that has a direction angle ($\alpha$) as a parameter to determine the accuracy depending on the distance from the boundaries of the correct decision. Thus, for an angle closer to the boundary of the referent human decision, the accuracy will be close to 1. In cases where the computer vision-based decision coincides with the referent human decision, the accuracy will be 1. Likewise, for all angles that deviate from the boundaries of the human decision by 15° or more degrees, the accuracy will be equal to 0. Since each simplified direction decision has a different referent range of angles, equations for accuracy are defined for all five cases.

The first case refers to determining the accuracy of the computer decision ($g_1$) based on a calculated direction angle ($\alpha$) for the simplified direction "straight" given by equation (1).

$$g_1(\alpha) = \begin{cases} e^{-(0.03*(|\alpha|-20)^2)}, & \alpha \in \langle-35,-20\rangle \cup \langle 20,35\rangle \\ 1, & \alpha \in [-20,20] \\ 0, & \alpha \in [-90,-35] \cup [35,90] \end{cases} \quad (1)$$

The accuracy of the decision for the direction "slight left" ($g_2$) is calculated according to equation (2).

$$g_2(\alpha) = \begin{cases} e^{-(0.03*(20-\alpha)^2)}, & \alpha \in \langle 5,20\rangle \\ e^{-(0.03*(\alpha-50)^2)}, & \alpha \in \langle 50,65\rangle \\ 1, & \alpha \in [20,50] \\ 0, & \alpha \in [-90,5] \cup [65,90] \end{cases} \quad (2)$$

Furthermore, the accuracy of the decision for the direction "slight right" ($g_3$) is calculated according to equation (3).

$$g_3(\alpha) = \begin{cases} e^{-(0.03*(|\alpha|-50)^2)}, & \alpha \in \langle-65,-50\rangle \\ e^{-(0.03*(20-|\alpha|)^2)}, & \alpha \in \langle-20,-5\rangle \\ 1, & \alpha \in [-50,-20] \\ 0, & \alpha \in [-90,-65] \cup [-5,90] \end{cases} \quad (3)$$

Equation (4) defines the accuracy of the decision for the direction "sharp left" ($g_4$).

$$g_4(\alpha) = \begin{cases} e^{-(0.03*(50-\alpha)^2)}, & \alpha \in \langle 35,50\rangle \\ 1, & \alpha \in [50,90] \\ 0, & \alpha \in [-90,35] \end{cases} \quad (4)$$

The last case is the accuracy for the direction "sharp right" ($g_5$) given by the equation (5).

$$g_5(\alpha) = \begin{cases} e^{-(0.03*(50-|\alpha|)^2)}, & \alpha \in \langle-50,-35\rangle \\ 1, & \alpha \in [-90,-50] \\ 0, & \alpha \in [-50,90] \end{cases} \quad (5)$$

Graphs in Fig. 5. show the relationship between the computer-determined direction angle and the accuracy of the decision made, which can take values from 0 to 1.

Once the criterion has been defined, it is possible to recalculate the average accuracy of the method for guiding. The method for guiding was again tested on a set of 500 images of crosswalks and 500 images of stairs. The evaluation results according to the new criterion have shown accuracy of 97.17% and 93.96% for the crosswalks and stairs respectively. Those results should be more realistic than those in Tab. 1 where rough assessment was made based only on a comparison of decisions with ground truth. However, due to lack of detailed explanations for evaluation methodology in similar papers, comparison with other approaches is not applicable.

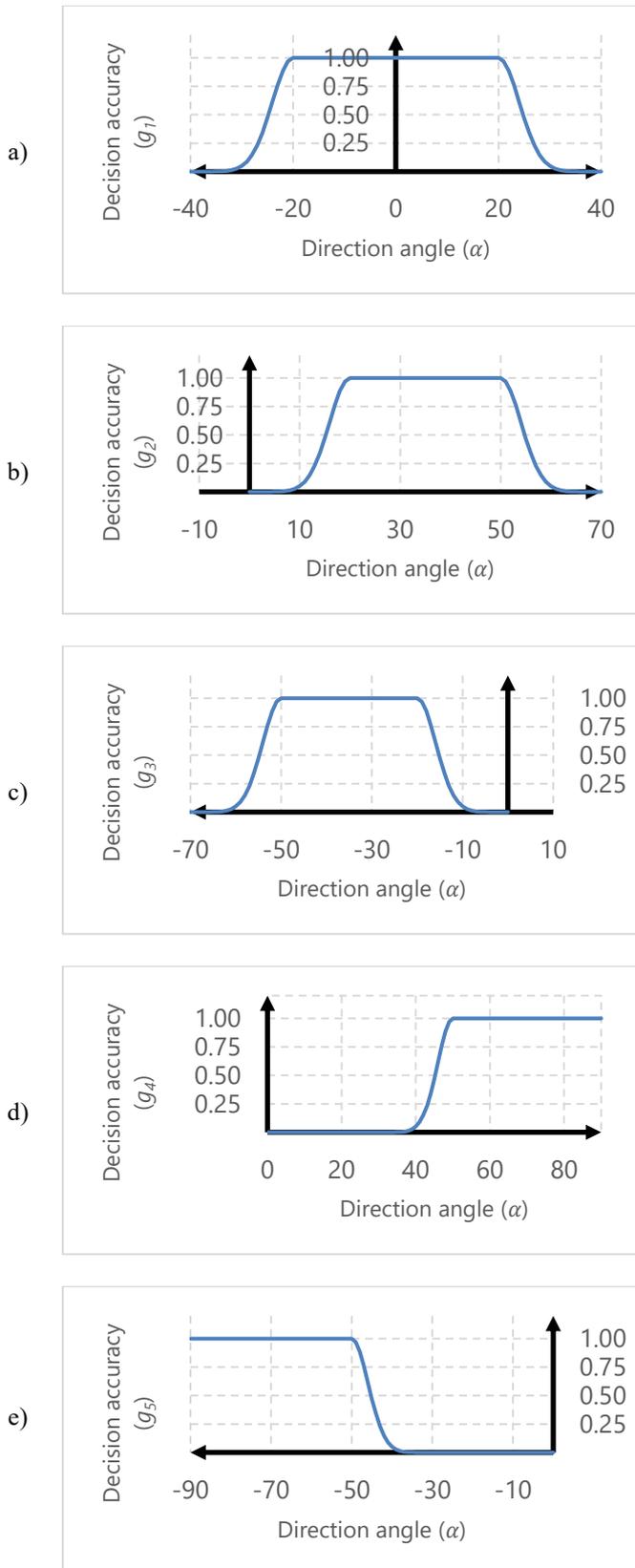

Figure 5. Criterion graphs for decisions: a) straight, b) slight left, c) slight right, d) sharp left, e) sharp right

## IV. Conclusion

Presented evaluation framework generally facilitates the testing process of computer vision methods for guiding the visually impaired persons. Except basic comparison with the ground truth, the framework includes the new criterion for assessment of calculated direction angles when guiding the visually impaired users. This is primarily important when evaluating the simplified direction instructions based on a direction angle. Although the proposed criterion is presented and tested on the example of guidance on stairs and pedestrian crosswalks, it can be concluded that it is universally applicable to other guidance procedures. Since proposed criterion and all parameters were obtained empirically, future work regarding this framework will incorporate machine learning techniques for dealing with direction angles and simplified directions.

### Acknowledgment

This work has been supported in part by the Croatian Science Foundation under the project UIP-2017-05-4968.